\documentclass{article}

    \usepackage[final]{neurips_2022}

\usepackage[utf8]{inputenc} 
\usepackage[T1]{fontenc}    
\usepackage{hyperref}       
\usepackage{url}            
\usepackage{booktabs}       
\usepackage{amsfonts}       
\usepackage{nicefrac}       
\usepackage{microtype}      
\usepackage{xcolor}         
\usepackage{mathtools}
\usepackage{wrapfig}
\usepackage{amssymb}
\usepackage{amsthm}

\DeclareMathOperator*{\argmax}{arg\,max}

\title{Denoising Deep Generative Models}

\author{%
  Gabriel Loaiza-Ganem \\
  Layer 6 AI\\
  \texttt{gabriel@layer6.ai}\\
   \And
  Brendan Leigh Ross \\
  Layer 6 AI\\
  \texttt{brendan@layer6.ai}\\
   \And
  Luhuan Wu \\
  Columbia University\\
  \texttt{lw2827@columbia.edu}\\
  \AND
  John P. Cunningham \\
  Columbia University\\
  \texttt{jpc2181@columbia.edu}\\
     \And
  Jesse C. Cresswell \\
  Layer 6 AI\\
  \texttt{jesse@layer6.ai}\\
     \And
  Anthony L. Caterini \\
  Layer 6 AI\\
  \texttt{anthony@layer6.ai}\\
}

\begin{document}

\maketitle

\begin{abstract}
Likelihood-based deep generative models have recently been shown to exhibit pathological behaviour under the manifold hypothesis as a consequence of using high-dimensional densities to model data with low-dimensional structure. In this paper we propose two methodologies aimed at addressing this problem. Both are based on adding Gaussian noise to the data to remove the dimensionality mismatch during training, and both provide a denoising mechanism whose goal is to sample from the model as though no noise had been added to the data. 
Our first approach is based on Tweedie's formula, and the second on models which take the variance of added noise as a conditional input. We show that surprisingly, while well motivated, these approaches only sporadically improve performance over not adding noise, and that other methods of addressing the dimensionality mismatch are more empirically adequate.
\end{abstract}

\section{Introduction}\label{sec:intro}
The manifold hypothesis \citep{bengio2013representation}, which states that high-dimensional data often lies on an unknown low-dimensional manifold embedded in ambient space, aims to explain the success of deep learning: neural networks would be unable to learn good low-dimensional representations if there was no low-dimensional structure to begin with. There have also been empirical studies estimating the intrinsic dimension of commonly-used image datasets, finding it is indeed much lower than its corresponding ambient dimension \citep{pope2020intrinsic, tempczyk2022lidl, brown2022union}. Along with these empirical verifications of the low-dimensional structure present in data, there has also been in a surge in research in deep generative models (DGMs) attempting to directly account for the manifold hypothesis \citep{gemici2016normalizing, dai2019diagnosing,
saremi2019neural,rezende2020normalizing, brehmer2020flows, mathieu2020riemannian, arbel2020generalized, kothari2021trumpets,  caterini2021rectangular, ross2021tractable, de2022riemannian, loaiza-ganem2022diagnosing, ross2022neural}. This is a relevant line of research, especially for likelihood-based models, which 
have been shown to suffer from \emph{manifold overfitting} under the manifold hypothesis \citep{dai2019diagnosing, loaiza-ganem2022diagnosing}, a surprising phenomenon where likelihoods can become arbitrarily large without recovering the ground truth distribution, even in the presence of an infinite amount of data.

Current DGMs that account for the manifold hypothesis require either non-trivial modifications from their corresponding fully-dimensional counterparts \citep{brehmer2020flows, arbel2020generalized, kothari2021trumpets,  caterini2021rectangular, ross2021tractable, ross2022neural}, or require training more models \citep{dai2019diagnosing, loaiza-ganem2022diagnosing}. In this paper we propose two slight modifications to existing full-dimensional likelihood-based models so as to enable them to directly account for the manifold hypothesis. In our first proposed method, we train off-the-shelf models on data to which Gaussian noise has been added -- so as to remove the dimensionality mismatch which causes manifold overfitting in the first place -- and then use Tweedie's formula \citep{robbins1956empirical} as a denoising step, i.e. as a correction to account for the fact that we have learned a noisy version of the target distribution rather than the ground truth distribution itself. In our second proposal, we also add Gaussian noise with variance $\sigma^2$ to the data, this time for a range of different values of $\sigma$. We then leverage conditional DGMs \citep{sohn2015learning, agrawal2016deep} to learn the conditional distribution of the (noisy) data given $\sigma$, and denoising is carried out by using $\sigma=0$ when sampling from the model.

In spite of being strongly motivated, both of our proposed procedures do not obtain consistent improvements over simply using full-dimensional models, unlike some of the aforementioned more involved manifold-aware models. We hope that this surprising result will lead into further research aiming to understand the interplay between the manifold hypothesis and DGMs.

\section{Background}
\subsection{Likelihood-based DGMs and Tweedie's formula}

Throughout this work we will assume that we have access to samples from a distribution $p(x)$ in $\mathbb{R}^D$. We will also assume that the manifold hypothesis holds; i.e., that $p(x)$ is supported on an embedded submanifold of $\mathbb{R}^D$ of dimension less than $D$.\footnote{While the notation $p(x)$ suggests this is a density in the Lebesgue sense, we highlight that formally $p$ is a probability measure as it is supported on a low-dimensional manifold. We nonetheless opt for this notation for consistency with most of the DGM literature.} Our discussions apply to all continuous likelihood-based models such as variational autoencoders (VAEs) \citep{kingma2013auto, rezende2014stochastic}, normalizing flows (NFs) \citep{dinh2016density, kingma2018glow, behrmann2019invertible, NEURIPS2019_5d0d5594, durkan2019neural, cornish2020relaxing}, energy-based models \citep{du2019implicit}, and continuous autoregressive models \citep{uria2013rnade, theis2015generative}, in which a density $p_\eta(x)$ over $\mathbb{R}^D$ is constructed through neural networks parameterized by $\eta$, and trained through maximum-likelihood
\begin{equation}\label{eq:ml}
    \eta^* = \argmax_\eta \mathbb{E}_{p(x)} [\log p_\eta(x)]
\end{equation}
with the intention of recovering $p(x)$. Note that we slightly abuse notation, as depending on the model being used, $p_\eta(x)$ might not be directly available. For example, VAEs maximize a lower bound of the log-likelihood, and energy-based models do not directly have access to $p_\eta(x)$ although they still aim to solve \eqref{eq:ml} through gradient estimates. We nonetheless keep the notation $p_\eta(x)$ for the sake of generality and provide a review of the DGMs that we will use in our experiments section in appendix \ref{app:dgms}. Likelihood-based DGMs do not properly account for the manifold hypothesis, since $p_\eta(x)$ is a high-dimensional density. As we will see in \autoref{sec:manifold_overfitting}, this modelling choice results in pathological behaviour, which we aim to address through Tweedie's formula and conditional DGMs.

\paragraph{Tweedie's formula} Assume we are given a sample $x_\sigma$, obtained by first sampling $x$ from $p(x)$, and then adding Gaussian noise $x_\sigma \coloneqq x + \sigma \epsilon$, where $\sigma > 0$ and $\epsilon \sim \mathcal{N}(0, I_D)$. Tweedie's formula \citep{robbins1956empirical} provides the best estimate $\hat{x}_\sigma$ (in mean squared error) of $x$ obtainable from $x_\sigma$:
\begin{equation}\label{eq:tweedie}
    \hat{x}_\sigma \coloneqq \mathbb{E}_{p(x|x_\sigma)}[x] = x_\sigma + \sigma^2 \nabla_{x_\sigma} \log p(x_\sigma).
\end{equation}
Surprisingly, computing $\hat{x}_\sigma$ does not require access to $p(x)$ nor to $p(x|x_\sigma)$, only the marginal of $x_\sigma$, $p(x_\sigma)$, is involved.

\paragraph{Conditional models} Many DGMs, including VAEs and NFs, 
admit conditional variants \citep{sohn2015learning, agrawal2016deep}. These models are trained to maximize
\begin{equation}\label{eq:cml}
    \eta^* = \argmax_\eta \mathbb{E}_{p(x,c)}[\log p_\eta(x|c)],
\end{equation}
where $c$ is a conditioning variable. For example, $c$ might be a class label, in which case $p(x,c)=p(x)p(c|x)$ where $p(c|x)$ is a point mass at the class label corresponding to $x$; or $c$ could also specify a subset of coordinates of $x$, in which case $p(c|x)$ selects the conditioning coordinates, often independently of $x$, i.e. $p(c|x) = p(c)$. Here $p_\eta(x|c)$ is now a density defined through neural networks parameterized by $\eta$, whose inputs now also include $c$. Once again, we include details of the conditional models that we use in appendix \ref{app:cdgms}.

\newpage

\subsection{Manifold overfitting}\label{sec:manifold_overfitting}

\begin{wrapfigure}[17]{r}{0.55\textwidth}
\vspace{-3.7em}
\begin{center}
\centerline{\includegraphics[width=0.75\linewidth]{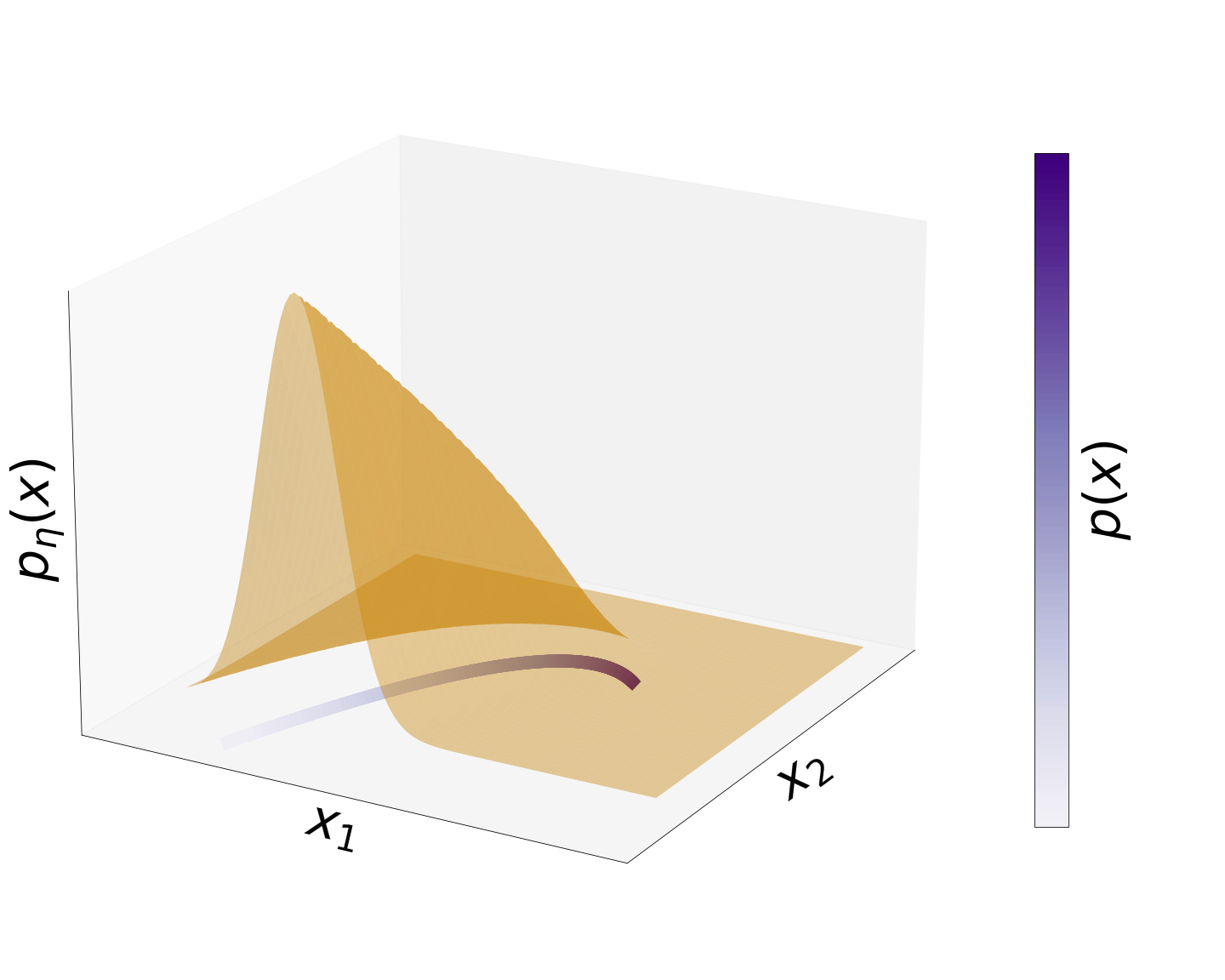}}
\caption{Illustration of manifold overfitting, where the ground truth distribution $p(x)$ in the $1$-dimensional curve (purple) is poorly approximated by the $2$-dimensional density $p_\eta(x)$ (orange), which nonetheless achieves large log-likelihoods $\mathbb{E}_{p(x)}[\log p_\eta(x)]$.}
\label{fig:manifold_overfitting}
\end{center}
\end{wrapfigure}
Manifold overfitting \citep{dai2019diagnosing, loaiza-ganem2022diagnosing} shows that solving \eqref{eq:ml} will in general not result in $p_{\eta^*}(x)$ recovering $p(x)$, as the likelihood $p_{\eta^*}(x)$ can achieve arbitrarily large values by concentrating around the manifold over which $p(x)$ is supported, without getting the correct distribution on the manifold. \autoref{fig:manifold_overfitting} illustrates this phenomenon.
Here $p(x)$ is supported on a $1$-dimensional curve (manifold) in $\mathbb{R}^2$, and the plotted choice of $p_\eta(x)$ concentrates around the correct manifold, but does so in an incorrect way, assigning more probability to the wrong region of the the curve. If $p_\eta(x)$ is flexible enough, this spiking behaviour can increase, resulting in unbounded likelihoods even if the model is not close to $p(x)$. Manifold overfitting strongly motivates the development of likelihood-based DGMs which properly account for the manifold hypothesis.

\section{Methods}\label{sec:methods}
\subsection{Tweedie Denoising DGMs}\label{sec:tweedie_dgms}

Here we propose to train a DGM not to learn $p(x)$ directly, but rather its noisy version $p(x_\sigma)$, which is the density obtained after convolving $p(x)$ with Gaussian noise with variance $\sigma^2$, where $\sigma$ is treated as a hyperparameter. This amounts to solving
\begin{equation}
    \eta^* = \argmax_\eta \mathbb{E}_{p(x_\sigma)} [\log p_\eta(x_\sigma)]
\end{equation}
instead of \eqref{eq:ml}. The intuition is simple: by adding noise, the target distribution $p(x_\sigma)$ is not supported on a low-dimensional manifold anymore (formally, it becomes absolutely continuous with respect to the Lebesgue measure in $\mathbb{R}^D$), which should theoretically avoid manifold overfitting. 
We point out that adding Gaussian noise is a common practice (see \autoref{sec:related}), although \citet{loaiza-ganem2022diagnosing} found that, by itself, doing so does not fully avoid manifold overfitting in practice. 
We thus propose to add an additional denoising step through Tweedie's formula in the hope of improving empirical performance by properly accounting for the fact that the learned distribution is not directly $p(x)$. Once we have the trained model $p_{\eta^*}(x_\sigma)$, given a sample $x_\sigma$ from the model, we correct it through Tweedie's formula \eqref{eq:tweedie}:
\begin{equation}\label{eq:tweedie_correction}
    \hat{x}_\sigma \leftarrow x_\sigma + \sigma^2 \nabla_{x_\sigma} \log p_{\eta^*}(x_\sigma).
\end{equation}
We highlight the simplicity of using Tweedie denoising DGMs: we only have to add Gaussian noise to training data, train an off-the-shelf likelihood-based DGM $p_\eta(x_\sigma)$, and do a post-hoc correction through \eqref{eq:tweedie_correction} at sample time.

\subsection{Conditional Denoising DGMs}

We also propose to use conditional models \eqref{eq:cml} to learn the conditional distribution of noisy data, conditional on the standard deviation of the added noise by maximizing
\begin{equation}
    \eta^* = \argmax_\eta \mathbb{E}_{p(x_\sigma|\sigma)p(\sigma)}[\log p_\eta(x_\sigma|\sigma)],
\end{equation}
where $p(\sigma)$ is an arbitrary distribution over $\sigma$, e.g. uniform on $(0, C)$ for some hyperparameter $C > 0$. Note that since we are now treating $\sigma$ as random instead of a fixed hyperparameter, we write $p(x_\sigma|\sigma)$ instead of $p(x_\sigma)$ for the distribution of noisy data at a given noise level. We also highlight the simplicity of using conditional denoising DGMs: during training we sample $\sigma$ along with each datapoint, add corresponding Gaussian noise to the data, and condition the DGM on $\sigma$. When sampling from a trained model, we simply sample from $p_{\eta^*}(x_\sigma|\sigma=0)$. The intuition is similar to that of Tweedie denoising DGMs: by adding noise, we hope the model manages to properly learn $p(x_\sigma|\sigma)$ for every $\sigma$ in the appropriate range, e.g. $(0, C)$, and that the $\sigma=0$ case $p(x)=p(x_\sigma|\sigma=0)$ is also learned by ``continuity'' over $\sigma$.

\section{Related work}\label{sec:related}
As mentioned in the introduction, most deep generative modelling methods that account for the manifold hypothesis without explicitly adding noise to the data deviate substantially from their full-dimensional counterparts. While not in itself a problem, this property does prevent ``plugging in'' any likelihood-based DGM from the vast existing literature to our context of interest. \citet{dai2019diagnosing} and \citet{loaiza-ganem2022diagnosing} propose to first obtain low-dimensional representations of the data, and then train a likelihood-based DGM on these representations, which results in the added complexity of having to specify two models and not having a single end-to-end training procedure. Indeed, our original motivation was to tackle the same problem in a simpler way.

Adding noise to data before training DGMs is a common practice \citep{vincent2008extracting, vincent2011connection, alain2014regularized, theis2015note, chae2021likelihood}, albeit not always directly motivated as a way to account for the manifold hypothesis. In the context of accounting for the manifold hypothesis within likelihood-based DGMs, several methods based on adding noise have been proposed, although these tend to be model-specific. For example, the method of \citet{zhang2020spread} can only be applied to VAEs, those of \citet{horvat2021denoising,
horvat2021density} and \citet{cunningham2021} to NFs, and \citet{meng2021improved} apply theirs only to autoregressive models. \citet{song2019generative} follow a similar approach for score-based models \citep{hyvarinen2005estimation}. Similarly, Tweedie's formula has been used in the context of DGMs before \citep{saremi2019neural, meng2021estimating}, although these uses are once again model-specific. \citet{kadkhodaie2021stochastic} apply Tweedie's formula iteratively  with the help of a denoiser \citep{mohan2020robust} and obtain strong empirical results, although their method turns the denoiser into a DGM, rather than fix manifold overfitting for existing DGMs. Finally, \citet{kim2020softflow} propose conditional denoising in the context of normalizing flows, and show that it works well in low-dimensional data: our contribution here can be understood both as highlighting the methodological generality of conditional denoising, and also in our empirical experiments on higher dimensions in the next section. Our motivation for this paper was to propose a widely applicable methodology, compatible with any likelihood-based DGM. 

We were also motivated by diffusion models, which have extremely strong empirical performance. These models can be understood as likelihood-based models \citep{ho2020denoising, song2021maximum}, or as score-based models in a stochastic differential equation setting \citep{song2021scorebased}. 
Diffusion models learn how to slowly transform noisy samples into samples from the data distribution (i.e., to denoise them). In our notation this roughly translates to transforming samples $x_{\sigma_2}$ from $p(x_{\sigma_2}|\sigma_2)$ into samples $x_{\sigma_1}$ from $p(x_{\sigma_1}|\sigma_1)$ for a multitude of values $\sigma_1 < \sigma_2$.
Importantly, the structure of diffusion models implies that these models learn not only the target distribution $p(x)$, but also noisy versions of it $p(x_\sigma|\sigma)$ at different noise levels $\sigma$. Furthermore, in contrast to likelihood-based models which can experience manifold overfitting, diffusion models are known to converge under the manifold hypothesis \citep{pidstrigach2022score, de2022convergence}. All these properties of diffusion models motivated our conditional models, with the hope that learning $p(x_\sigma|\sigma)$ for a continuum of values of $\sigma$ could address manifold overfitting.

\section{Experiments}\label{sec:experiments}
\subsection{Results}

Although as mentioned previously our methods can in principle be applied to any likelihood-based model, in this section we focus on VAEs and NFs as both have commonly-used conditional versions. For example, energy-based models \citep{du2019implicit} keep a sample buffer during training, and na\"{i}vely adding the conditioning variable as input to the energy function would result in the buffer containing samples at different noise levels: this would confound any observed poor performance of conditional denoising, and attempting to improve upon the buffer falls outside of the scope of this work. Similarly, we also omit autoregressive models from our experiments, as most well-performing versions of these models are discrete rather than continuous \citep{van2016pixel, salimans2017pixelcnn++} (and are thus not susceptible to manifold overfitting), and proposing performant continuous autoregressive models also falls outside the scope of our work. We use the prefixes ``ND-'', ``TD-'', and ``CD-'' to denote models trained with added Gaussian noise with no denoising step, Tweedie denoising, and conditional denoising, respectively.  All experimental details are provided in appendix \ref{app:exps}, and our code is publicly available at \url{https://github.com/layer6ai-labs/denoising_dgms}.
\begin{wraptable}{r}{0.55\textwidth}
\vspace{-0.3em}
 \caption{FID scores (lower is better). Means $\pm$ standard errors across $3$ runs are shown. Best models and those whose standard errors overlap with those of the best model are bolded.}
    \label{table:fid}
    \begin{center}
    \scalebox{0.7}{
    \begin{tabular}{lrrrr}
    \toprule
      MODEL   & MNIST & FMNIST & SVHN & CIFAR-10\\
     \midrule
     VAE & $\mathbf{197.4\pm 1.5}$ & $\mathbf{188.9 \pm 1.8}$ & $ 311.5 \pm 6.9 $ & $ 270.3 \pm 3.2 $ \\
     ND-VAE & $199.9 \pm 1.4$ & $\mathbf{185.7 \pm 2.0}$ & $317.8 \pm 8.3$ & $264.5 \pm 0.5$ \\
     TD-VAE & $199.1 \pm 0.8 $ & $\mathbf{190.4 \pm 3.3}$ & $ 310.9 \pm 8.9 $ & $263.9 \pm 0.9$ \\
     CD-VAE & $\mathbf{197.4 \pm 0.2} $ & $195.8 \pm 2.1$ & $\mathbf{290.0 \pm 4.4}$ & $ \mathbf{262.4 \pm 0.3}$ \\
     \midrule
     NF & $137.2 \pm 3.4$ & $110.5 \pm 0.9$ & $ \mathbf{231.9 \pm 22.0} $ & $\mathbf{222.7 \pm 3.9} $ \\
     ND-NF & $103.2 \pm 0.4 $ & $72.3 \pm 0.8$ & $ 222.0 \pm 5.7 $ & $ \mathbf{222.9 \pm 1.2} $ \\
     TD-NF & $105.6 \pm 0.5 $ & $\mathbf{70.6 \pm 0.4}$ & $ 224.2 \pm 4.4 $ & $ \mathbf{222.8 \pm 2.2} $ \\
     CD-NF & $\mathbf{87.4 \pm 0.5}$ & $73.3 \pm 0.3$ & $\mathbf{206.0 \pm 7.1}$ & $\mathbf{225.4 \pm 0.7}$ \\
     \bottomrule
    \end{tabular}
    }
    \end{center}
\end{wraptable}

\autoref{table:fid} shows comparisons of all the considered models using the FID score \citep{heusel2017gans} for the MNIST \citep{lecun1998mnist}, FMNIST \citep{xiao2017fashion}, SVHN \citep{netzer2011reading}, and CIFAR-10 \citep{krizhevsky2009learning} datasets. We opted to use FID scores as a measure of how well the models recover $p(x)$ instead of test log-likelihoods since the latter are, by definition, unable to detect manifold overfitting. We highlight that we tuned $\sigma$ for the Tweedie denoising models, as well as $C$ for the conditional denoising models. We can see that, surprisingly, the TD-VAE (TD-NF) and CD-VAE (CD-NF) models do not consistently outperform their VAE (NF) and ND-VAE (ND-NF) baselines: the only instances of denoising models obtaining a non-marginal improvement over their baselines are the CD-VAE on SVHN and the CD-NF on MNIST. We also tried annealing $\sigma$ (for Tweedie denoising models) and $C$ (for conditional denoising models), but found results did not significantly change. Not only do our denoising models not outperform simply adding Gaussian noise, but in some cases denoising can even hamper performance: for example CD-VAEs on FMNIST and TD-NFs on MNIST both significantly -- albeit marginally -- underperform their non-denoised alternatives. 

\subsection{Discussion}
We conjecture that the issue with these methods remains related to manifold overfitting: the target noisy distribution might still be very peaked around the manifold and difficult to learn, in which case $p_{\eta^*}(x_\sigma)$ (or $p_{\eta^*}(x_\sigma|\sigma)$ for conditional models) might not be close to its target $p(x_\sigma)$ (or $p(x_\sigma|\sigma)$) and just concentrate around the manifold. If this is the case, we should expect neither Tweedie's formula nor conditioning on $\sigma=0$ to properly sample from $p(x)$. In other words, the noise being added might not be enough to numerically overcome manifold overfitting. We hypothesize that using more powerful models and further tuning how the noise is added might alleviate the situation.

Another potential explanation for the performance of Tweedie denoising models is that the value of $\sigma$ that we used ($0.01$ for most models, which as previously mentioned was found by tuning this hyperparameter) is too small, and thus the update from \eqref{eq:tweedie_correction} ends up barely correcting the samples. We find this explanation is not fully satisfactory as due to manifold overfitting, one should expect $||\nabla_{x_\sigma} \log p_{\eta^*}(x_\sigma)||_2$ term to become larger as $\sigma$ becomes smaller (since the density should become ``spikier'' around the manifold), potentially offsetting the small correction size $\sigma$. Additionally, this explanation does not provide any intuition for the observed performance of our conditional models.

Yet another explanation would be that the manifold hypothesis does not hold, and that thus manifold overfitting does not happen to begin with. We find this hypothesis particularly hard to believe: first, the manifold hypothesis is a sensible and intuitive way to think about high-dimensional data \citep{bengio2013representation} that been empirically verified in various ways \citep{pope2020intrinsic, tempczyk2022lidl, brown2022union}. Second, the works of \citet{dai2019diagnosing} and \citet{loaiza-ganem2022diagnosing} show very clear empirical improvements by avoiding manifold overfitting.

Finally, it is also likely the case that observed data \emph{already} represents noisy observations from a true low-dimensional manifold, and that injecting further noise to accommodate our approaches makes the problem significantly harder at the outset. However, the relative good performance of the no denoising (ND) models over their vanilla versions without any added noise suggests that adding noise is not making the problem harder.

\section{Conclusions}\label{sec:conclusions}
In this paper we propose Tweedie denoising and conditional denoising with the goal of alleviating manifold overfitting. Our methodologies are based on adding Gaussian noise to the data before training a likelihood-based DGM, along with a way of denoising samples from the resulting trained models. Unexpectedly, our denoising approaches do not provide meaningful empirical improvements;
we suspect that manifold overfitting remains a culprit in the failure of these models.
We hope that our work will incentivize the community to further understand this intriguing result, as well as the role of the manifold hypothesis in DGMs.



{
\small
\bibliography{refs}
\bibliographystyle{abbrvnat}
}

\appendix

\section{Appendix}
\subsection{Likelihood-based DGMs}\label{app:dgms}

\paragraph{VAEs} Variational autoencoders \citep{kingma2013auto, rezende2014stochastic} model $x \in \mathbb{R}^D$ through a lower-dimensional latent variable $z \in \mathbb{R}^d$. A prior $p(z)$, often taken as a standard Gaussian, is specified, and the conditional distribution $p_\theta(x|z)$ is parameterized by a neural network with parameters $\theta$. In Gaussian VAEs, $p_\theta(x|z)$ (often referred to as the decoder) is given by a Gaussian whose parameters are given by the output of the neural network parameterized by $\theta$. Since the $x$-marginal of the model, $\int p(z)p_\theta(x|z) dz$, is not tractable, VAEs cannot be trained through maximum-likelihood directly. Instead, an auxiliary distribution $q_\phi(z|x)$ (the encoder) is introduced, with the aim of approximating the posterior $p_\theta(z|x)$. The approximate posterior $q_\phi(z|x)$ is often taken as a low-dimensional Gaussian whose parameters are given by the output of a neural network parameterized by $\phi$. A lower bound to the low-likelihood is then maximized over $\eta=(\theta, \phi)$:
\begin{equation}\label{eq:vaes}
    \eta^* = \argmax_\eta \mathbb{E}_{p(x)}\left[\mathbb{E}_{q_\phi(z|x)}[\log p_\theta(x|z)] - \mathbb{KL}(q_\phi(z|x) || p(z))\right].
\end{equation}

\paragraph{NFs} Normalizing flows \citep{dinh2016density} construct $p_\eta(x)$ as the density obtained by transforming $x=f_\eta(z)$, where $f_\eta: \mathbb{R}^D \rightarrow \mathbb{R}^D$ is a bijective neural network parameterized by $\eta$, and $z \sim p(z)$, where $p(z)$ is often taken as a standard Gaussian. By the change of variable formula, we have that
\begin{equation}\label{eq:change_of_variable}
    p_\eta(x) = p(z)\left|\det \dfrac{\partial f_\eta^{-1}}{\partial x}\right|,
\end{equation}
where $z = f_\eta^{-1}(x)$ and $\partial f_\eta^{-1} / \partial x$ is the Jacobian matrix of $f_\eta^{-1}$ evaluated at $x$. As long as $f_\eta$ is constructed in such a way that its inverse and the determinant in \eqref{eq:change_of_variable} can be efficiently computed, NFs can be trained through maximum-likelihood \eqref{eq:ml}. In practice, $f_\eta^{-1}$ is constructed (rather than $f_\eta$) by stacking coupling layers. Each coupling layer is itself a bijective transformation with the aforementioned properties enabling tractability. Coupling layers proceed by partitioning their input into two blocks $x=(x_A, x_B)$, and the corresponding output $z=(z_A, z_B)$ is given by:
\begin{equation}\label{eq:coupling}
    \begin{cases}
    z_A = x_A\\
    z_B = g(x_B; h_\eta(x_A)),
    \end{cases}
\end{equation}
where $g(\cdot;h):\mathbb{R} \rightarrow \mathbb{R}$ is an invertible function parameterized by $h$ which is applied element-wise to $x_B$, and $h_\eta$ is a neural network mapping $x_A$ to the parameters of $g$. For example, $g$ could be an affine transformation \citep{dinh2016density} parameterized by two scalars $h \in \mathbb{R}^2$, or a monotonic rational quadratic spline \citep{durkan2019neural}. It is easy to check that the determinant of a coupling layer is triangular (up to a permutation), and can thus be computed efficiently. Finally, $f_\eta^{-1}$ is constructed by stacking multiple coupling layers, each with its own partition, on top of each other.

\subsection{Conditional likelihood-based DGMs}\label{app:cdgms}

\paragraph{VAEs} Variational autoencoders can straightforwardly be made into conditional models \citep{sohn2015learning} by modifying \eqref{eq:vaes} as follows:
\begin{equation}
    \eta^* = \argmax_\eta \mathbb{E}_{p(x,c)}\left[\mathbb{E}_{q_\phi(z|x,c)}[\log p_\theta(x|z)] - \mathbb{KL}(q_\phi(z|x,c) || p(z|c))\right],
\end{equation}
where the prior $p(z|c)$ can now also depend on the conditioning variable $c$, although this dependency can be omitted, i.e. $p(z|c) = p(z)$. In practice, this change amounts to having the encoder and decoder neural networks take $c$ as input in addition to $x$ and $z$, respectively. We omitted the dependency of $p(z|c)$ in our experiments for the sake of simplicity, although in some preliminary experiments we did not observe any significant changes by including this dependency.

\paragraph{NFs} Similarly to VAEs, NFs can be made into conditional models \citep{agrawal2016deep} defining a conditional density $p_\eta(x|c)$ simply by using the conditioning variable $c$ as an input to coupling layers, i.e. modifying \eqref{eq:coupling} to
\begin{equation}\label{eq:c_coupling}
    \begin{cases}
    z_A = x_A\\
    z_B = g(x_B; h_\eta(x_A, c)),
    \end{cases}
\end{equation}
which then enables the model to be trained through \eqref{eq:cml}. This strategy has been used in different contexts \citep{atanov2019semi, ardizzone2019guided, winkler2019learning}.

\subsection{Experimental details}\label{app:exps}

For all experiments, we used the Adam optimizer \citep{kingma2014adam}, and gradient clipping with a value of $10$. For ND- models, we tuned $\sigma$ by reporting the best value out of $\sigma \in \{0.005, 0.01, 0.05, 0.1, 0.5\}$. We use the same value of $\sigma$ for the TD- versions of the models. For the CD- models, we similarly tune $C$ by selecting the best value in $\{0.005, 0.01, 0.05, 0.1, 0.5\}$.

\paragraph{VAEs} We preprocessed the data by scaling it to $[0,1]$. The latent space is $20$-dimensional, the learning rate is $0.001$, and we train the models for $100$ epochs. We use fully-connected architectures for MNIST and FMNIST, both for the encoder and decoder, which both have a single hidden layer with $256$ units. For SVHN and CIFAR-10, we use convolutional encoders and decoders, which have $(32, 32, 16, 16)$ and $(16, 16, 32, 32)$ hidden channels, respectively, with a fully-connected layer at the end of the encoder, and one at the beginning of the decoder. We use ReLU activations throughout. For the conditional versions on MNIST and FMNIST, we found that simply concatenating the $1$-dimensional $\sigma$ with a $784$-dimensional datapoint $x$ resulted in the network just ignoring $\sigma$. In order to provide the inductive bias that the conditioning is a relevant feature to the model, we used two additional networks, which we call conditioning networks. The encoder conditioning network takes $\sigma$ as input and outputs a $64$ dimensional representation, which is then concatenated with $x$ before being fed into the encoder. Similarly, for the decoder, the conditioning network takes $\sigma$ as input and outputs an $8$-dimensional representation, which is concatenated with $z$ before being fed into the decoder. Both conditioning networks have a single hidden layer with $256$ units and use ReLU activations. The conditioning network for the encoder on SVHN and CIFAR-10 remains the same, although the conditioning network for the encoder now outputs a $32\times32$ channel, which is concatenated with the input $x$ before being fed to the encoder.

\paragraph{NFs} We also preprocessed the data by scaling it to $[0,1]$ for MNIST and FMNIST, and by whitening it for SVHN and CIFAR-10. We used a learning rate of $0.0005$, and train for $100$ epochs with early stopping on validation log-likelihood with a patience of $30$ epochs. We use rational quadratic spline flows \citep{durkan2019neural} with $128$ units, $4$ layers, and $3$ blocks per layer. For the conditional models, as in VAEs, we use a conditioning network which takes $\sigma$ as input and outputs a $64$-dimensional representation, which is used as additional input in the coupling layers \eqref{eq:c_coupling}. The conditioning network has a single hidden layer with $256$ units and uses a ReLU activation.

\end{document}